\pdfoutput=1

\documentclass[11pt]{article}

\usepackage[preprint]{acl}

\usepackage{times}
\usepackage{latexsym}

\usepackage[T1]{fontenc}

\usepackage[utf8]{inputenc}

\usepackage{microtype}

\usepackage{inconsolata}

\usepackage{graphicx}
\usepackage{amssymb}
\usepackage{multirow}

\usepackage{dsfont}
\usepackage{multirow}
\usepackage{tabularx}
\usepackage{booktabs}
\usepackage{subfigure}

\usepackage{caption} 
\title{Mitigating Strategy Preference Bias in Emotional Support Conversation via Uncertainty Estimations}

\author{Yougen Zhou\textsuperscript{1}, Qin Chen\textsuperscript{2\thanks{Corresponding Author}}, Ningning Zhou\textsuperscript{3},  Jie Zhou\textsuperscript{2}, Xingjiao Wu\textsuperscript{4}, Liang He\textsuperscript{1, 2}\\
\textsuperscript{1}{Shanghai Institute of Artificial Intelligence for Education}, \textsuperscript{2}School of Computer Science and Technology, \\ \textsuperscript{3}School of Psychology and Cognitive Science, \textsuperscript{4}School of Pharmacy,\\ East China Normal University, Shanghai, China
}

\begin{document}
\maketitle
\begin{abstract}
Emotional support conversation (ESC) aims to alleviate distress through empathetic dialogue, yet large language models (LLMs) face persistent challenges in delivering effective ESC due to low accuracy in strategy planning. Moreover, there is a considerable preference bias towards specific strategies. Prior methods using fine-tuned strategy planners have shown potential in reducing such bias, while the underlying causes of the preference bias in LLMs have not well been studied. To address these issues, we first reveal the fundamental causes of the bias by identifying the knowledge boundaries of LLMs in strategy planning. Then, we propose an approach to mitigate the bias by reinforcement learning with a dual reward function, which optimizes strategy planning via both accuracy and entropy-based confidence for each region according to the knowledge boundaries. Experiments on the ESCov and ExTES datasets with multiple LLM backbones show that our approach outperforms the baselines, confirming the effectiveness of our approach.
\end{abstract}

\section{Introduction}

Emotional support conversation (ESC) aims to reduce distress through empathetic dialogue, supporting individuals facing personal challenges \cite{langford1997social, greene2003handbook, heaney2008social}. Effective ESC not only requires nuanced strategy selection to generate contextually appropriate responses but also avoids low-quality responses that could lead to unintended psychological issues \cite{burleson2003emotional}. In recent years, large language models (LLMs), with their advanced conversational abilities, have been increasingly integrated into various dialogue systems \cite{jia2023knowledge, lee2023prompted}. There is growing interest in harnessing LLMs for emotional support and professional counseling \cite{chen2023controllable, zheng2023building}.

\begin{figure}[t]
    \centering
    \includegraphics[width=0.95\linewidth]{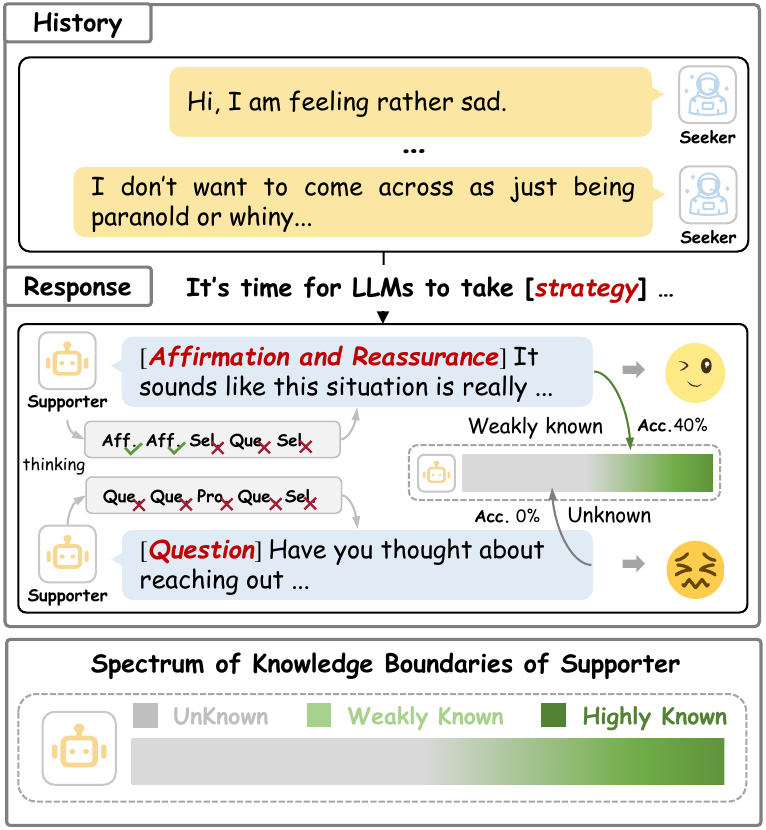}
    \caption{Examples of an LLM-based supporter with knowledge boundaries in emotional support. Weakly known areas reflect partial knowledge; unknown areas lie outside its correct knowledge boundaries.}
    \label{fig:figure_1}
\end{figure}
ESC tasks involve both strategy selection and strategy-constrained response generation, where choosing the appropriate strategy is critical for effective emotional support \cite{liu2021towards}. Despite their remarkable capabilities, LLMs often struggle to provide high-quality ESC due to two key limitations \cite{friedman2023leveraging, chen2023soulchat}. First, LLMs exhibit low accuracy in strategy selection, frequently failing to identify contextually appropriate support strategies \cite{zhao2023transesc, farhat2024chatgpt}. Second, LLMs display a severe strategy preference bias, rigidly favoring certain general or familiar strategies over others, which limits the adaptability to users' evolving emotional needs in different situations \cite{kang2024can, zhao2025chain}. To address these issues, previous methods have introduced fine-tuned strategy planners to reduce bias in both closed- and open-source LLMs \cite{kang2024can}, or applied reinforcement learning (RL) with external preference datasets to improve strategy selection \cite{zhao2025chain}. Whereas, these methods rely on additional modules or external data to mitigate preference bias, and the underlying causes of strategy preference bias in LLMs have not well been studied.

Recent research has begun to investigate the limitations of LLMs by examining the internal knowledge distributions learned during pretraining, which offers a new perspective to understand the origins of the preference bias. Previous studies show that the knowledge boundaries of LLMs are often ambiguous, encompassing substantial weakly known information \cite{gekhman2024does} (depicted as the light green area in the spectrum, Figure \ref{fig:figure_1}). As a result, LLMs may struggle to confidently generate accurate response even when relevant information is present for the weakly known areas \cite{zhang2024self}. In contrast, in the unknown areas out of the knowledge scope of LLMs (represented by the gray area in the spectrum, Figure \ref{fig:figure_1}), they usually exhibit overconfidence on the fabricated content \cite{zhang2024r}. Furthermore, as LLMs incrementally learn new knowledge through unknown samples, their susceptibility to producing ungrounded content tends to increase \cite{gekhman2024does}. Inspired by these findings, we posit that preference bias in ESC is not merely due to the data imbalance during the fine-tuning stage, but it is rooted in the internal knowledge distributions of LLMs in pretraining. As shown in Figure \ref{fig:figure_1}, the preference bias of strategies reflects a confidence calibration based on the strength of underlying knowledge. Therefore, we aim to explicitly identify the knowledge boundaries of LLMs for ESC, and leverage this internal structure to mitigate strategy preference bias.

In this paper, we propose an approach that leverages uncertainty estimation to explicitly model the knowledge boundaries of LLMs for ESC, which improves strategy planning by mitigating the preference bias. We first identify and characterize the distinct regions within LLMs’ knowledge distributions. Building on this, we employ reinforcement learning following supervised fine-tuning (SFT) with a dual reward function combining strategy accuracy and entropy confidence measures. This enables the model to optimize strategy planning across varying levels of knowledge certainty, prioritizing grounded responses while avoiding excessive dependence on overused strategies. Our approach is adaptable to various LLM backbones. Experiments on the ESCov and ExTES datasets demonstrate that our approach effectively improves the performance of LLMs in strategy planning for ESC. The main contributions of this paper are as follows:
\begin{itemize}
    \item We introduce a novel perspective by identifying the knowledge boundaries of LLMs in strategy planning for ESC, and uncover the pivotal role of weakly known samples in mitigating the strategy preference bias.
    \item We propose an approach that models different knowledge regions with different rewards, which balances well between the accuracy and preference in strategy planning.
    \item We conduct elaborate analyses of the experimental results on two benchmark datasets, and provide a better understanding of the effectiveness of our approach.
\end{itemize}

\section{Related Work}

\subsection{Emotional Support Conversation} 
Emotional support conversation (ESC) involves interactions between a seeker experiencing emotional distress and a supporter aiming to alleviate it \cite{liu2021towards}. Early approaches include global-to-local hierarchical graph networks to model dialogue context \cite{peng2022control}, incorporating commonsense knowledge for empathetic responses \cite{tu2022misc}, and modeling emotions and semantics to enhance response relevance \cite{zhao2023transesc}. With the rise of LLMs, recent work leverages their conversational capabilities and apply SFT the LLM with ESC task specialized dialogues, which outperforms general-purpose LLMs \cite{chen2023soulchat, qiu2024smile}. However, previous studies have suggested that they often struggle with selecting the correct strategy and a notable preference for a specific strategy \cite{kang2024can}. To make LLMs better emotional supporting, \cite{kang2024can} use fine-tuned strategy planners reduce bias in both closed- and open-source LLMs and \cite{zhao2025chain} optimize strategy selection with external preference dataset. Unlike these methods, our approach do not require any additional plugs or data to improve LLMs’ performance in strategy selection and response generation.

\subsection{Knowledge Boundary}
Knowledge boundaries of LLMs is critical for identifying their limitations in accurately expressing factual knowledge that they learned from the pre-training stage. While models are equipped with extensive parametric knowledge, some studies indicate their inability to discern the knowledge they possess from what they lack, thus failing to articulate their knowledge boundary \cite{yin-etal-2023-large, ren-etal-2025-investigating}. To enhance a model’s awareness of its knowledge boundary, prior work focus two areas: utilizing inherent knowledge to reduce the ratio of the model’s ``Unknown Knows” \cite{wei2022chain, li2023inference, tian2023fine} and acknowledging the knowledge it lacks to minimize the ratio of the model’s ``Unknown Unknows” \cite{zhang2024r, yang2024alignment}. \cite{gekhman2024does} find that LLMs struggle to acquire new factual knowledge through fine-tuning and that LLMs linearly increase the model’s tendency to hallucinate by learning the examples with new knowledge. Focused on this aspect, our work investigates how to enable models to express the area of knows consistently, while also considering exploration uknows in knowledge boundaries.

\subsection{Uncertainty Estimation for LLMs}
Uncertainty estimation aims to derive calibrated confidence scores from LLMs, reflecting their knowledge uncertainty in tasks \cite{manakul-etal-2023-selfcheckgpt, duan2024shifting}. These methods are categorized into four classes. Likelihood-based methods quantify sentence uncertainty using token probabilities \cite{vazhentsev-etal-2023-efficient}. Prompting-based methods instruct LLMs to express uncertainty verbally \cite{lin-etal-2022-truthfulqa, xiong2023can} or self-evaluate their correctness probability \cite{kadavath2022language}. Sampling-based methods measure response consistency \cite{xiong2023can} or semantic entropy \cite{kuhn2023semantic}. Training-based methods fine-tune LLMs to improve calibrated uncertainty expressions \cite{lin-etal-2022-truthfulqa}. However, both self-verbalized and sampling methods using extrinsic prompting or aggregation strategies need to concern themselves with detecting hallucinations, such as by setting uncertainty thresholds. Our work concentrates on enabling the model to explicitly express its response based on the knowledge boundary autonomously, rather than generating a probability score.

\begin{figure*}[t]
    \centering
    \includegraphics[width=0.95\linewidth]{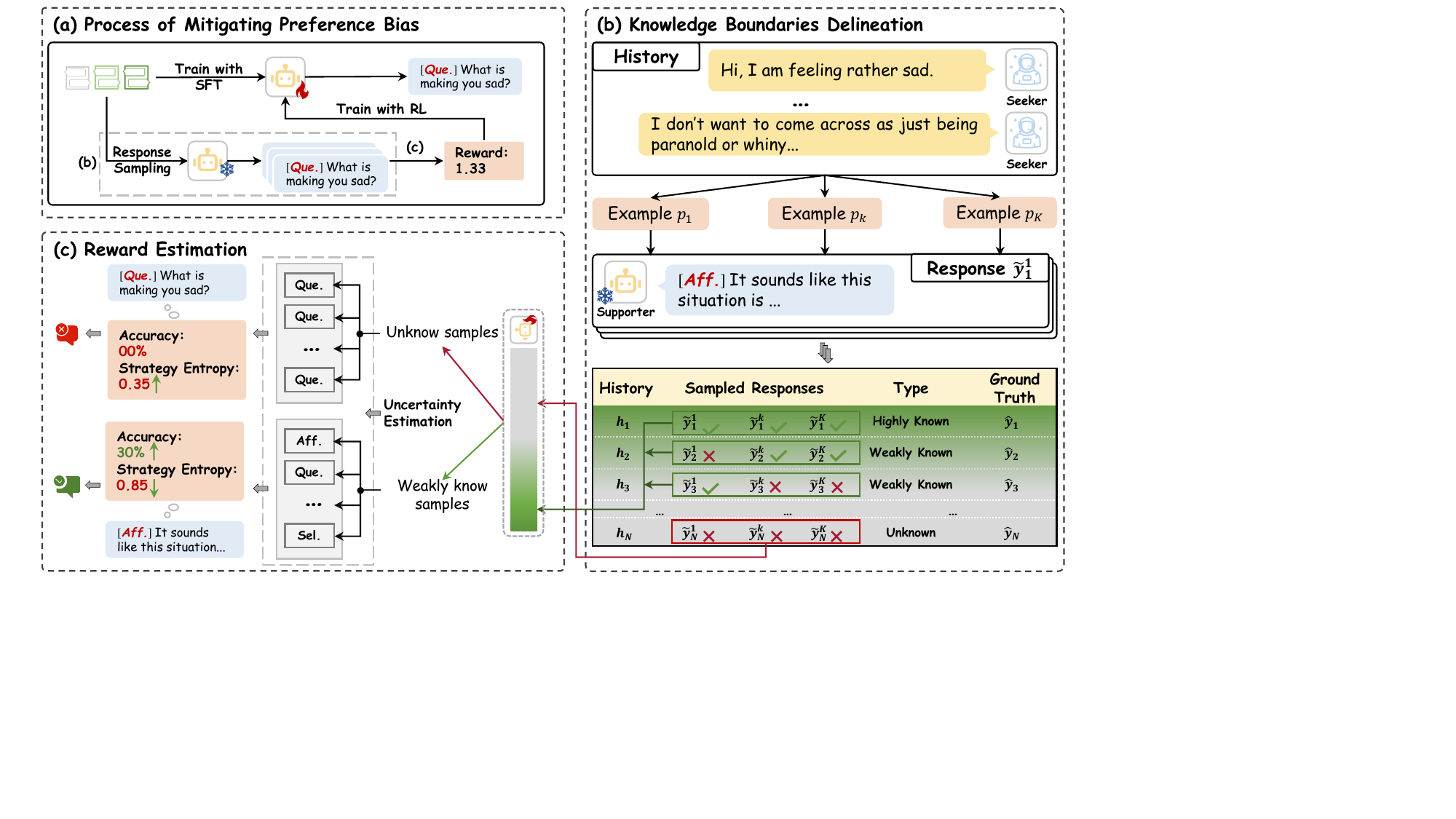}
    \caption{Overview of our approach. (a) illustrates the SFT and RL process for a pre-trained LLM to enable strategy selection and empathetic response generation for ESC. (b) shows the delineation of knowledge boundaries into highly known, weakly known, and unknown knowledge areas. (c) presents the calculation of reward score, combining strategy selection accuracy and entropy confidence, tailored to distinct sample types relative to knowledge boundaries.}
    \label{fig:figure_2}
\end{figure*}
\section{Method}
Figure \ref{fig:figure_2} illustrates the overall procedure of our proposed approach. The framework for mitigating strategy preference bias is shown in Figure \ref{fig:figure_2}(a), which consists of two stages. In the first stage, we perform supervised fine-tuning (SFT) on a pre-trained LLM using the full dataset. In the second stage, reinforcement learning (RL) is applied to the finetuned model using different samples categorized according to the model's knowledge boundaries. The process of knowledge boundaries delineation is depicted in Figure \ref{fig:figure_2}(b), we generate multiple responses for each input by repeating sampling from the model, and compute rewards based on both strategy accuracy and strategy entropy. Details about the ESC task and our method are presented as follows.

\subsection{Problem Formulation}
Following prior work \cite{liu2021towards}, the effectiveness of ESC responses generated by LLMs largely depends on the selection of an appropriate support strategy. We formalize the ESC task as a generation problem for a pre-trained LLM $\mathcal{M}$, where the model directly produces a response that incorporates the intended support strategy. Given a flattened dialogue history $h$, the model generates a response $\tilde{y}$ consisting of a special strategy$s$ followed by the actual utterance $y$, formalized as $\tilde{y} = s \oplus y \sim P(\tilde{y} \mid h;\theta_{\mathcal{M}})$, where $s$ corresponds to a predefined strategy in $S = \{s_1, s_2, \cdots, s_n\}$, and $\theta_{\mathcal{M}}$ denotes the parameters of the model. This formulation allows the model to implicitly select and express a strategy as part of the response generation process. As ESC is a strategy-centric task, \cite{kang2024can} suggest that a good supporter LLM should satisfy two key properties: \textbf{Proficiency}, the ability to select appropriate strategies; and \textbf{Preference}, the ability to avoid overusing specific strategies.

\subsection{Mitigating Preference Bias}
\subsubsection{Supervised Fine-tuning} SFT is a widely used approach to adapt LLMs to specific tasks. To improve the proficiency of the LLM $\mathcal{M}$ in ESC, we first perform SFT using a curated dataset $\mathcal{D} = \{(h_i, \tilde{y}_i)\}_{i=1}^{N}$ consisting of dialogue history-response pairs. To ensure the model clearly understands the task, we prepend both a task description and a strategy description as a part of the prompt. Additionally, beyond the dialogue history, we include dialogue background, which encompasses the seeker's problem, emotion, and situation gathered from a pre-chat survey. The model is finetuned by minimizing the negative log-likelihood:
\begin{equation}
    \mathcal{L}_{\mathrm{SFT}} = - \sum_{i=1}^{N} \log P_{\theta_{\mathcal{M}}}(\tilde{y}_i \mid h_i).
\end{equation}
This objective encourages the model to generate responses containing appropriate emotional support strategies.

However, there are some examples that introduce new knowledge for the LLM, and LLMs struggle to acquire new knowledge through fine-tuning, as fine-tuning examples that introduce new knowledge are learned significantly lower than those consistent with the model's knowledge. Moreover, as the examples with new knowledge are learned, it becomes increasingly prone to generating off-strategy responses. In the following subsections, we present our RL approach incorporating strategy accuracy-based and entropy-based rewards to reduce this limitation.

\subsubsection{Reinforcement Learning} Inspired by recent insights on knowledge boundaries, we propose adapting dynamic rewards to different regions of the model's knowledge boundary. Unlike prior methods that apply a uniform reward function across all samples, our approach directs the model toward distinct learning objectives based on samples falling within different knowledge regions.

Our goal is to guild the LLM's generation toward desirable responses by maximizing an alignment objective $\mathcal{R}$, which can take various forms such as format measures. Formally, we aim to maximize the expected reward:
\begin{equation}
    \mathds{E}_{h \sim \mathcal{D}, \tilde{y} \sim P_{\theta_{\mathcal{M}}}(\cdot|h)}[\mathcal{R}(h,\tilde{y})].
\end{equation}
We formulate this as an RL problem. We use the fine-tuned model to initialize the policy network as $\pi_0 = P_{\theta_{\mathcal{M}}}$ and update it using group relative policy optimization (GRPO). The token generation process for a response $\tilde{y}$ can be seen as a Markov Decision Process (MDP) $\langle \mathcal{S}, \mathcal{A}, r, \mathcal{P} \rangle$ with a state space $\mathcal{S}$, action space $\mathcal{A}$, reward function $r$, and state-transition probability $\mathcal{P}$. At each time step $t$, the agent (LLM) selects the next token from the vocabulary $\mathcal{V}$ based on the current policy network $\pi(\tilde{y}|h,\tilde{y}_{<t})$. The episode ends upon generation an end-of-sequence token, producing the complete response $\tilde{y}$. We can fine-tune the policy network $\pi$ by maximizing the expected reward $r$:
\begin{equation}
    \mathds{E}_{\pi}[r]=\mathds{E}_{h \sim \mathcal{D},\tilde{y} \sim \pi (\cdot|h)}[r(h,\tilde{y})].
\end{equation}

Recall that our goal is to maximize the proficiency of the model $\mathcal{M}$ and minimize its preference bias. For each input dialogue history $h_i$, we seek high confidence correct predictions, while also allowing for controlled exploration via alternative responses. To this end, we propose a reward estimation approach that dynamically adjusts the reward function $r$ according to the model's estimated knowledge boundary for each input $h_i$.

\subsection{Knowledge Boundaries Delineation}
To categorize samples relative to the model's knowledge boundary, we define three regions: \textit{highly known}, \textit{weakly known}, and \textit{unknown}. As illustrated in Figure \ref{fig:figure_2}(b), we estimate the location of each dialogue history $h_i$ within this boundary by analyzing the model’s output probabilities, assigning each example to one of the three regions.

\subsubsection{Response Sampling} To capture the variability in the model’s behavior, we generate multiple responses for each dialogue history. Given a dataset $\mathcal{D} = \{(h_i, \hat{y}_i)\}_{i=1}^{N}$, where $\hat{y}_i=s_i\oplus y_i$ denotes the ground-truth strategy $s_i$ and corresponding response content $y_i$, we employ few-shot prompting with a temperature of $T = 0.4$ to balance response accuracy and diversity while mitigating prompt sensitivity. The prompt set $P$ includes $K$ distinct one-shot examples to facilitate in-context learning.
 
For each dialogue history $h_i$, the $k$-th sampling iteration combines a few-shot example $p_k \in P$ with $h_i$ to prompt the model $\mathcal{M}$, producing the $k$-th response $\tilde{y}_i^k$. Repeating this process $K$ times yields a response set $Y_i = \{\tilde{y}_i^k\}_{k=1}^K$. Corresponding labels $Z_i = \{z_i^k\}_{k=1}^K$ are assigned by comparing each generated strategy within response $\tilde{y}_i^k$ with the ground-truth $\hat{y}_i$, where $z_i^k \in {0, 1}$ (1 indicates a correct response, and 0 indicates an incorrect one). This results in a data tuple $(h_i, Y_i, Z_i, \hat{y}_i)$ to calculate response accuracy subsequently.

\subsubsection{Deriving Knowledge Categories} To assess the model’s proficiency for each dialogue history $h_i$, we compute an accuracy-based confidence score $c_i$, defined as the proportion of responses with the correct strategy:
\begin{equation}
    c_i = \frac{1}{K} \sum_{k=1}^{K} \mathds{1}(s_i^k = s_i),
\end{equation}
where $\mathds{1}(\cdot)$ is the indicator function. Based on $c_i$, we categorize each example as follows:
\begin{itemize}
    \item $c_i = 1$: Highly known samples, where all generated strategies are correct.
    \item $c_i = 0$: Unknown samples, where no generated strategies are correct.
    \item $0 < c_i < 1$: Weakly known samples, exhibiting partial correctness in strategy predictions.
\end{itemize}

\subsection{Reward Estimation} The confidence score $c_i$ serves as a reward signal to encourage the model toward higher strategy selection proficiency. To further promote consistency, we incorporate an entropy-based metric that quantifies the diversity of strategies expressed across the sampled responses. Specifically, we estimate the empirical distribution over strategies by computing the relative frequency of each strategy $s \in S$:
\begin{equation}
    p(s \mid h_i) = \frac{1}{K} \sum_{k=1}^K \mathds{1}(s_i^k = s),
\end{equation}
where $\mathds{1}(\cdot)$ is the indicator function. The corresponding strategy entropy $e_i$ is calculated as:
\begin{equation}
    e_i = - \sum_{s \in S} p(s \mid h_i)\log p(s \mid h_i),
\end{equation}
which measures the degree of strategy variation in the model's outputs. To tailor reward optimization to different regions of the model’s knowledge boundary, we define region-specific reward functions.

For \textit{highly known} and \textit{weakly known} samples, where the model exhibits full or partial proficiency, we prioritize consistency by a reward that emphasizes low diversity:
\begin{equation}
    r_{known}(h_i, \tilde{y}_i) = 1 - \frac{e_i}{\log(|S|)},
\end{equation}
where $|S|$ is the number of predefined strategies. This encourages the model to converge toward the correct strategy with minimal variability. 

Conversely, for \textit{unknown} samples, where the model lacks proficiency, we encourage exploration to discover potentially correct responses. The reward function for these samples promotes high diversity:
\begin{equation}
    r_{unknow}(h_i, \tilde{y}_i) = \frac{e_i}{\log(|S|)},
\end{equation}
which incentivizes the generation of a wider range of strategies that may lead to discovery of effective responses.

To keep the policy network $\pi$ from moving too far from the initial model $P_{\mathcal{M}}$, we also add a KL-divergence penalty reward. Therefore, the final reward becomes:
\begin{equation}
    r(h_i, \tilde{y}_i) = c_i + r_{region}(h_i, \tilde{y}_i) -\beta \log\frac{\pi(\tilde{y}_i|h_i)}{P_{\theta_{\mathcal{M}}}(\tilde{y}_i|h_i)},
\end{equation}
where $r_{region}$ is either $r_{known}$ or $r_{unknow}$ based on the sample's classification, and the hyperparameter $\beta$ is the coefficient of KL-penalty. This composite reward structure ensures that the model balances proficiency, consistency, and exploration while maintaining stability during optimization.

\section{Experimental Setup}
\subsection{Datasets}
We conduct experiments on two emotional support datasets: ESCov \cite{liu2021towards} and ExTES \cite{zheng-etal-2024-self}. \textbf{ESCov} collected through crowdworkers acting as help-seekers and supporters, categorizes emotional support strategies into 8 distinct types. \textbf{ExTES} generated iteratively using ChatGPT, leverages dialogue examples from authoritative emotional support datasets, which classify strategies into 16 types. For ESCov, we utilize the training set introduced by \cite{liu2021towards} for model training and knowledge boundary delineation. For evaluation, we employ the test set proposed by \cite{kang2024can}. For ExTES, we adopt an 8:2 split for training and test sets, respectively, and follow the test set construction methodology outlined in \cite{kang2024can} for the test set to ensure consistency and comparability.

\subsection{Evaluation Metrics} 
As described in prior studies \cite{kang2024can}, a good supporter LLM should satisfy two properties: Proficiency and Preference. For proficiency, we focus on strategy selection accuracy, measured by \textbf{macro F1} score $\mathcal{Q}$ and \textbf{weighted F1} score $\mathcal{Q}_{\mathcal{W}}$. For preference, we select \textbf{strategy preference bias} $\mathcal{B}$ as metric, which quantifies the deviation between the model’s selected strategies and the ideal strategy preferences. Additionally, \textbf{ROUGE-L} (R-L) is also adopted to evaluate the semantic aspects of the generated response.

\subsection{Baselines} 
To evaluate the effectiveness of our approach, we benchmark it against several baselines that designed to mitigate preference biases in LLMs using self-contact techniques. We implement three self-contact methods: (1) \textit{Direct-Refine}, where the LLM iteratively refines its initial response; (2) \textit{Self-Refine} \cite{madaan2023self}, which improves the initial response through self-generated feedback; and (3) \textit{Few-Shots}, which samples some example from training sets to guide response generation. Given the prevalence of SFT in enhancing ESC capabilities, we also compare against full-parameters SFT models. Additionally, we evaluate a baseline using GRPO \cite{shao2024deepseekmath} with rewards based on response format and correctness to isolate the impact of general RL tuning from our region-aware reward design.

\subsection{Implementation Details} 
We implement our method on LLaMA-3.1-8B-Instruct \cite{grattafiori2024llama} and Qwen2.5-7B-Instruct \cite{team2024qwen2}. For SFT, we train the model via LLaMA-Factory \cite{zheng2024llamafactory} with full parameters. The input format follows the official chat templates, while the output is the target response consisting of strategy. The models are trained for 2 epochs with a learning rate of $1 \times e^{-5}$. For RL, we train the fine-tuned model via VERL \cite{sheng2024hybridflow} for 300 episodes, a batch size of 256, and a learning rate of $1 \times e^{-6}$. The KL-penalty coefficient is set to 0.001. All experiments are conducted on 2 NVIDIA Tesla A800 GPUs. We set maximum input length to 1024 tokens and maximum target length of 256 tokens across all backbones.

\section{Results and Analyses}
\begin{table*}[!th]
\centering
\begin{tabular}{l c c c c c c c c c}
\toprule
\multirow{2}{*}{\textbf{Method}} & \multicolumn{4}{c}{\textbf{ESCov}} & & \multicolumn{4}{c}{\textbf{ExTES}} \\
\cline{2-5} \cline{7-10}
 & $\mathcal{Q}$ $\uparrow$ & $\mathcal{B}$ $\downarrow$ & $\mathcal{Q}_{\mathcal{W}}$ $\uparrow$ & \textbf{R-L} $\uparrow$ & & $\mathcal{Q}$ $\uparrow$ & $\mathcal{B}$ $\downarrow$ & $\mathcal{Q}_{\mathcal{W}}$ $\uparrow$ & \textbf{R-L} $\uparrow$\\
\midrule
\multicolumn{10}{c}{\textbf{Qwen-2.5-7B-Instruct}} \\
Standard Prompting & 8.13 & 2.22 & 8.49 & 13.56 & & 13.57 & 3.87 & 24.74 & 20.15 \\
Few-Shot (2-shot) & 12.41 & 1.19 & 13.84 & 13.47 & & 11.78 & 3.83 & 20.75 & 20.41 \\
Direct-Refine & 12.51 & 1.22 & 14.55 & 13.78 & & 13.31 & 3.86 & 24.38 & 19.75 \\
Self-Refine & 13.36 & \underline{1.16} & 15.38 & 13.54 & & 11.81 & 3.86 & 20.98 & 19.75 \\
SFT & 22.10 & 1.79 & 22.80 & 19.07 & & 40.48 & 1.06 & 57.74 & 27.78 \\
GRPO & \textbf{26.26} & 1.73 & \underline{26.46} & \underline{19.12} & & \underline{45.96} & \underline{0.65} & \underline{61.27} & \underline{28.62} \\
\textbf{Ours} & \underline{26.16} & \textbf{1.12} & \textbf{28.24} & \textbf{19.86} & & \textbf{47.51} & \textbf{0.64} & \textbf{63.79} & \textbf{29.60} \\
\midrule 
\multicolumn{10}{c}{\textbf{LLaMA-3.1-8B-Instruct}} \\
Standard Prompting & 7.59 & 2.34 & 7.96 & 11.76 & & 6.84 & 3.85 & 11.42 & 16.26 \\
Few-Shot (2-shot) & 12.61 & \textbf{0.81} & 12.00 & 12.98 & & 11.98 & 3.81 & 19.29 & 19.52 \\
Direct-Refine & 9.45 & 1.69 & 10.04 & 12.41 & & 6.75 & 3.85 & 11.51 & 16.82 \\
Self-Refine & 10.87 & 1.36 & 11.89 & 12.97 & & 6.89 & 3.86 & 11.02 & 17.79 \\
SFT & 21.79 & 1.30 & 23.67 & 15.70 & & 41.34 & 0.83 & 57.22 & 36.06 \\
GRPO & \underline{22.69} & 1.27 & \underline{25.74} & \underline{15.79} & & \underline{45.83} & \underline{0.75} & \underline{60.04} & \underline{37.69} \\
\textbf{Ours} & \textbf{24.27} & \underline{1.03} & \textbf{27.37} & \textbf{15.84} & & \textbf{48.79} & \textbf{0.73} & \textbf{64.74} & \textbf{39.84} \\
\bottomrule
\end{tabular}
\caption{Automatic evaluation results on the ESCov and ExTES datasets using Qwen-2.5-7B-Instruct and LLaMA-3.1-8B-Instruct as backbones. $\uparrow$ indicates higher is better, $\downarrow$ indicates lower is better. The best results are highlighted in \textbf{bold}. The second-best results are \underline{underlined}.}
\label{tab:1}
\end{table*}

\subsection{Main Results}
Table \ref{tab:1} presents the performance comparison between our proposed method and several baselines. Our approach consistently improves strategy selection accuracy while effectively reducing strategy preference bias across all evaluated models. Specifically, while SFT and GRPO without explicit preference optimization enhance strategy selection accuracy, they simultaneously increase strategy preference bias, limiting their adaptability to seeker's diverse emotional needs. In contrast, our approach mitigates preference bias while enhancing accuracy, underscoring the importance of incorporating preference-aware optimization in ESC tasks. Moreover, Our method also enhances ESC performance across all backbone models, regardless of their initial capabilities. It enhances strategy accuracy and reduces bias in both Qwen2.5-7B-Instruct and LLaMA-3.1-8B-Instruct, demonstrating its versatility and robustness.

Our results also highlight the limitations of self-contact methods, which cannot achieve a steady increase in strategy accuracy. Although their strategy bias is low, they cannot accurately select the right strategy. This is slightly different from the conclusions of previous work \cite{kang2024can}, which have shown that self-contact methods can cause increased policy bias and decreased policy accuracy.

\begin{table}[!t]
\centering
\resizebox{\linewidth}{!}{
\begin{tabular}{lcccc}
\toprule
\textbf{Ours} vs. \textbf{GRPO} & Win & Lose & Tie & $k$ \\
\midrule
Acceptance & 41.33 & 26.00 & 32.66 & 0.611 \\
Effectiveness & 38.67 & 27.33 & 34.00 & 0.650 \\
Sensitivity & 40.00 & 26.00 & 34.00 & 0.669 \\
Satisfaction & 41.33 & 24.67 & 34.00 & 0.729 \\
\bottomrule
\end{tabular}
}

\caption{Human evaluation results comparing Ours and GRPO. Win, Lose, and Tie percentages reflect cases where our method, GRPO, or neither were preferred, respectively. Cohen’s kappa ($k$) indicates inter-annotator agreement.}
\label{tab:human_eval}
\end{table}
\subsection{Human Evaluation Results}
Table \ref{tab:human_eval} presents the human evaluation results comparing our method with GRPO on 50 dialogue samples from the ESCov dataset. Three annotators rated each response pair across four criteria: \textit{Acceptance}, \textit{Effectiveness}, \textit{Sensitivity}, and \textit{Satisfaction}. Our method consistently outperforms GRPO, achieving higher win rates across all dimensions, ranging from 38.67\% to 41.33\%. These results suggest that our responses are generally perceived as more satisfying by human evaluators. The inter-annotator agreement ranges from 0.611 to 0.729, indicating substantial agreement among annotators. This reinforces the robustness of our method in improving emotional support quality.

\begin{figure*}[ht]
\centering
\subfigure[Performance comparison of different backbones using various samples of knowledge boundaries from the ESCov dataset.]{
\centering
\includegraphics[width=\textwidth]{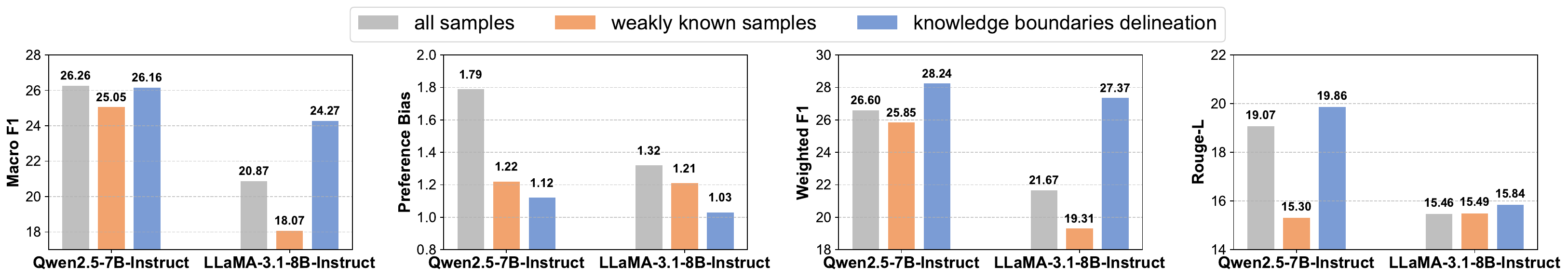} 
\label{fig:ablation_1}
}
\subfigure[Performance of reward functions varying different uses of uncertainty measures on the ESCov dataset.]{
\centering
\includegraphics[width=\textwidth]{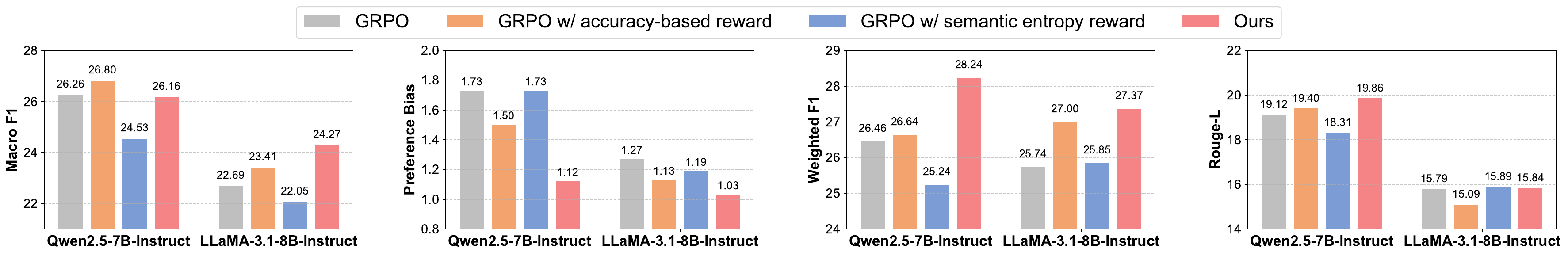} 
\label{fig:ablation_2}
}
\caption{Effects of knowledge boundary delineation and uncertainty estimation rewards on ESCov.}
\label{fig:overall}
\end{figure*}
\subsection{Ablation Study}
\subsubsection{Effect of Knowledge Boundary Delineation}
We conduct an ablation study to assess the effectiveness of knowledge boundary delineation. Specifically, we compare three configurations: (1) our full method with dynamic rewards based on knowledge regions, (2) a baseline without boundary awareness, and (3) a variant that trains on weakly known samples using uniform rewards, simulating a coarse partition. All models are trained via SFT on the ESCov dataset and then optimized using RL with identical hyperparameters. As illustrated in Figure~\ref{fig:ablation_1}, our approach outperforms both variants across all backbones, achieving higher strategy accuracy and lower preference bias. Interestingly, training solely on weakly known samples reduces preference bias more effectively than the baseline, highlighting their value in reducing model overconfidence. However, this variant still underperforms our full method, likely due to its inability to leverage the exploratory potential of unknown-region examples. These results underscore the value of fine-grained knowledge boundary delineation in guiding reward design and improving alignment efficiency in ESC.

\subsubsection{Effect of Uncertainty Estimation Reward}
We further conduct an ablation study to evaluate the impact of our two uncertainty-based reward components during the RL stage. Specifically, we compare: (1) a baseline using standard format and correctness-based rewards, (2) an accuracy-based reward that reflects the model’s confidence alignment with ground-truth labels, and (3) a entropy-based reward that quantifies strategy diversity via the entropy of the predicted strategy distribution. All models are fine-tuned using SFT on the ESCov dataset, followed by RL under identical settings. As shown in Figure~\ref{fig:ablation_2}, the accuracy-based reward consistently improves performance across all evaluation metrics for both LLMs. In contrast, using entropy alone leads to less stable performance, sometimes degrading specific metrics. Notably, combining both accuracy and entropy achieves the best overall performance, demonstrating the complementary nature of the two signals. These results validate the effectiveness of our dual uncertainty-guided reward design in enhancing strategy accuracy and reducing bias in ESC.

\begin{table}[!t]
\centering
\resizebox{\linewidth}{!}{
\begin{tabular}{lcccc}
\toprule
\textbf{Data Type} & $\mathcal{Q}$ $\uparrow$ & $\mathcal{B}$ $\downarrow$ & $\mathcal{Q}_{\mathcal{W}}$ $\uparrow$ & \textbf{R-L} $\uparrow$ \\
\midrule
\midrule
\multicolumn{5}{c}{\textbf{Qwen-2.5-7B-Intruct}} \\
\midrule
Full Dataset & 22.10 & 1.79 & 22.80 & 19.07 \\
Full Dataset(Ours) & 26.16 & 1.12 & 28.24 & 19.86 \\
Weakly Knows(Ours) & 20.91 & 1.22 & 22.19 & 15.30 \\
Weakly Knows(\textit{Prompt}) & 18.16 & 1.35 & 20.45 & 15.12 \\
\midrule
\multicolumn{5}{c}{\textbf{LLaMA-3.1-8B-Intruct}} \\
\midrule
Full Dataset & 21.79 & 1.30 & 23.67 & 15.70 \\
Full Dataset(Ours) & 24.27 & 1.03 & 27.37 & 15.84 \\
Weakly Knows(Ours) & 18.07 & 1.11 & 19.31 & 15.89 \\
Weakly Knows(\textit{Prompt}) & 17.19 & 1.27 & 19.30 & 15.50 \\
\bottomrule
\end{tabular}
}

\caption{Performance metrics across different data types. \textit{Prompt} refers a prompt-based uncertainty estimation.}
\label{tab:tab_3}
\end{table}
\subsection{Analysis}
\subsubsection{Effect of Weakly Known Samples}
We analyze the effect of training on weakly known samples identified via uncertainty estimation. Specifically, we compare four configurations: (1) full dataset SFT, (2) our method applied to the full dataset, (3) SFT with only weakly known samples selected by our accuracy based estimator, and (4) a prompt based method where weakly known samples are self-reported by the model. As shown in Table~\ref{tab:tab_3}, using weakly known samples consistently reduces strategy preference bias across all backbones. These findings indicate that weakly known samples can mitigate bias effectively. However, this reduction in bias comes at the cost of strategy proficiency. This trade-off likely stems from the reduced quantity of training examples when filtering for only weakly known samples. Notably, our accuracy-based method outperforms the prompt-based approach in both proficiency and bias reduction, demonstrating the reliability of our uncertainty estimation. 

\begin{figure}[!t]
    \centering
    \includegraphics[width=0.95\linewidth]{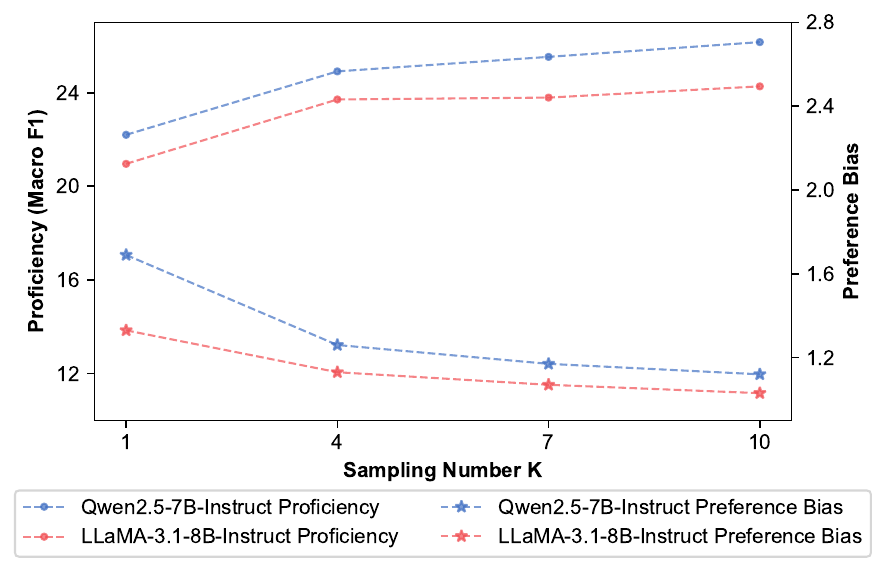}
    \caption{Experiments of proficiency and preference bias of various sampling number on ESCov.}
    \label{fig:analysis_2}
\end{figure}
\subsubsection{Effect of Sampling Number}
The sampling number $K$ is a key hyper-parameter, significantly impacting the precision of knowledge boundary measurements. We evaluated its effects by testing $K$ values, with results shown in Figure \ref{fig:analysis_2} for the ESCov dataset. Experiments reveal that increasing $K$ from small values markedly enhances proficiency and reduces preference bias. However, beyond a certain point, performance gains plateau, indicating convergence. As larger $K$ values incur substantial computational costs, we avoided testing beyond $K=10$. Setting $K=10$ effectively balances performance improvements with computational efficiency, closely approximating true knowledge boundaries.

\section{Conclusion} 
In this paper, we propose an approach to identify the knowledge boundaries of LLMs for the ESC task, and explicitly model different regions with a dual reward function combining strategy accuracy and entropy confidence measures for strategy planning. Extensive experiments on ESCov and ExTES datasets show that our approach outperforms the state-of-the-art baselines in both proficiency and bias reduction. In particular, the results underscore the importance of weakly known samples to enhance preference bias mitigation. In the future, we will extend our approach to other dialogue tasks, and refine knowledge boundary definitions to adapt to different downstream tasks.

\section*{Limitations}

In real counseling sessions, each session typically lasts around 45 minutes, with about five sessions focusing on specific topics. While our dataset involves longer multi-turn interactions compared to others, it remains significantly shorter than actual counseling sessions. Future work should aim to include longer conversations and multi-session interactions to better emulate counseling scenarios.

\bibliography{custom}

\appendix

\end{document}